\documentclass[twoside,11pt]{article}
\usepackage{haldefs}
\usepackage{algorithm}
\usepackage{algorithmic}
\usepackage{psfig}
\usepackage{theapa}
\usepackage{jair}
\usepackage{url}

\jairheading{26}{2006}{101-126}{8/05}{5/06}
\ShortHeadings{Domain Adaptation for Statistical Classifiers}{Daum\'e III \& Marcu}
\firstpageno{101}

\newcommand{\D}{\mathcal{D}}
\newcommand{\megam}{\textsc{Mega} Model}
\newcommand{\system}[1]{\textsc{#1}}

\newcommand{\pinch}{\vspace{-1mm}}

\newcommand{\mention}[3]{\underline{#1}${}_{\textsc{#2}}^{\textsc{#3}}$}

\newcommand{\din}{^{\textsf{(i)}}}
\newcommand{\dou}{^{\textsf{(o)}}}
\newcommand{\dgg}{^{\textsf{(g)}}}
\newcommand{\dnu}{^{\textsf{(}\nu\textsf{)}}}

\begin{document}

\title{Domain Adaptation for Statistical Classifiers}

\author{\name Hal Daum\'e III \email hdaume@isi.edu \\
        \name Daniel Marcu    \email marcu@isi.edu \\
        \addr Information Sciences Institute\\
              University of Southern California\\
              4676 Admiralty Way, Suite 1001\\
              Marina del Rey, CA 90292 USA}

\maketitle

\begin{abstract}%
The most basic assumption used in statistical learning theory is that
training data and test data are drawn from the same underlying
distribution.  Unfortunately, in many applications, the ``in-domain''
test data is drawn from a distribution that is related, but not
identical, to the ``out-of-domain'' distribution of the training data.
We consider the common case in which labeled out-of-domain data is
plentiful, but labeled in-domain data is scarce.  We introduce a
statistical formulation of this problem in terms of a simple mixture
model and present an instantiation of this framework to maximum
entropy classifiers and their linear chain counterparts.  We present
efficient inference algorithms for this special case based on the
technique of conditional expectation maximization.  Our experimental
results show that our approach leads to improved performance on three
real world tasks on four different data sets from the natural language
processing domain.
\end{abstract}

\section{Introduction}

The generalization properties of most current statistical learning
techniques are predicated on the assumption that the training data and
test data come from the same underlying probability distribution.
Unfortunately, in many applications, this assumption is inaccurate.
It is often the case that plentiful labeled data exists in one domain
(or coming from one distribution), but one desires a statistical model
that performs well on another related, but not identical domain.  Hand
labeling data in the new domain is a costly enterprise, and one often
wishes to be able to leverage the original, ``out-of-domain'' data
when building a model for the new, ``in-domain'' data.  We do not seek
to \emph{eliminate} the annotation of in-domain data, but instead seek
to minimize the amount of new annotation effort required to achieve
good performance.  This problem is known both as \emph{domain
adaptation} and \emph{transfer}.

In this paper, we present a novel framework for understanding the
domain adaptation problem.  The key idea in our framework is to treat
the in-domain data as drawn from a mixture of two distributions: a
``truly in-domain'' distribution and a ``general domain''
distribution.  Similarly, the out-of-domain data is treated as if
drawn from a mixture of a ``truly out-of-domain'' distribution and a
``general domain'' distribution.  We apply this framework in the
context of conditional classification models and conditional
linear-chain sequence labeling models, for which inference may be
efficiently solved using the technique of conditional expectation
maximization.  We apply our model to four data sets with varying
degrees of divergence between the ``in-domain'' and ``out-of-domain''
data and obtain predictive accuracies higher than any of a large
number of baseline systems and a second model proposed in the
literature for this problem.

The domain adaptation problem arises very frequently in the natural
language processing domain, in which millions of dollars have been
spent annotating text resources for morphological, syntactic and
semantic information.  However, most of these resources are based on
text from the news domain (in most cases, the Wall Street Journal).
The sort of language that appears in text from the Wall Street Journal
is highly specialized and is, in most circumstances, a poor match to
other domains.  For instance, there has been a recent surge of
interest in performing summarization \cite{elhadad05medical} or
information extraction \cite{hobbs02ie} of biomedical texts,
summarization of electronic mail \cite{rambow04email}, information
extraction from transcriptions of meetings, conversations or voice-mail
\cite{huang01voicemail}, among others.  Conversely, in the machine
translation domain, most of the parallel resources that machine
translation system depend on for parameter estimation are drawn from
transcripts of political meetings, yet the translation systems are
often targeted at news data \cite{munteanu05comparable}.


\section{Statistical Domain Adaptation} \label{sec:adaptation}

In the multiclass classification problem, one typically assumes the
existence of a training set $\D = \{ (x_n,y_n) \in \mathcal{X} \times
\mathcal{Y} : 1 \leq n \leq N \}$, where $\mathcal{X}$ is the input
space and $\mathcal{Y}$ is a finite set.  It is assumed that each
$(x_n,y_n)$ is drawn from a fixed, but unknown base distribution $p$
and that the training set is independent and identically distributed,
given $p$.  The learning problem is to find a function $f :
\mathcal{X} \fto \mathcal{Y}$ that obtains high predictive accuracy
(this is typically done either by explicitly minimizing the
regularized empirical error, or by maximizing the probabilities of the
model parameters).

\subsection{Domain Adaptation}

In the context of domain adaptation, the situation becomes more
complicated.  We assume that we are given \emph{two} sets of training
data, $\D\dou$ and $\D\din$, the ``out-of-domain'' and ``in-domain''
data sets, respectively.  We no longer assume that there is a single
fixed, but known distribution from which these are drawn, but rather
assume that $\D\dou$ is drawn from a distribution $p\dou$ and $\D\din$
is drawn from a distribution $p\din$.  The learning problem is to find
a function $f$ that obtains high predictive accuracy on data drawn
from $p\din$.  (Indeed, our model will turn out to be symmetric with
respect to $\D\din$ and $\D\dou$, but in the contexts we consider
obtaining a good predictive model of $\D\din$ makes more intuitive
sense.)  We will assume that $|\D\dou| = N\dou$ and $|\D\din| =
N\din$, where typically we have $N\din \ll N\dou$.  As before, we will
assume that the $N\dou$ out-of-domain data points are drawn iid from
$p\dou$ and that the $N\din$ in-domain data points are drawn iid from
$p\din$.

Obtaining a good adaptation model requires the careful modeling of the
relationship between $p\din$ and $p\dou$.  If these two distributions
are independent (in the obvious intuitive sense), then the
out-of-domain data $\D\dou$ is useless for building a model of $p\din$
and we may as well ignore it.  On the other hand, if $p\din$ and
$p\dou$ are identical, then there is no adaptation necessary and we
can simply use a standard learning algorithm.  In practical problems,
though, $p\din$ and $p\dou$ are neither identical nor independent.

\subsection{Prior Work} \label{prior-work}

There has been relatively little prior work on this problem, and
nearly all of it has focused on specific problem domains, such as
n-gram language models or generative syntactic parsing models.  The
standard approach used is to treat the out-of-domain data as ``prior
knowledge'' and then to estimate maximum a posterior values for the
model parameters under this prior distribution.  This approach has
been applied successfully to language modeling
\cite{roark03lmadaptation} and parsing \cite{roark03adaptation}.  Also
in the parsing domain, \citeA{hwa99induction} and
\citeA{gildea01variation} have shown that simple techniques based on
using carefully chosen subsets of the data and parameter pruning can
improve the performance of an adapted parser.  These models assume a
data distribution $\p{\cD \| \th}$ with parameters $\th$ and a prior
distribution over these parameters $\p{\th \| \et}$ with
hyper-parameters $\et$.  They estimate the $\et$ hyperparameters from
the out-of-domain data and then find the maximum a posteriori
parameters for the in-domain data, with the prior fixed.

In the context of conditional and discriminative models, the only
domain adaptation work of which we are aware is the model of
\citeA{chelba04adapt}.  This model again uses the out-of-domain data
to estimate a prior distribution, but does so in the context of a
maximum entropy model.  Specifically, a maximum entropy model is
trained on the out-of-domain data, yielding optimal weights for that
problem.  These weights are then used as the \emph{mean} weights for
the Gaussian prior on the learned weights for the in-domain data.

Though effective experimentally, the practice of estimating a prior
distribution from out-of-domain data and fixing it for the estimation
of in-domain data leaves much to be desired.  Theoretically, it is
strange to estimate and \emph{fix} a prior distribution from data;
this is made more apparent by considering the form of these models.
Denoting the in-domain data and parameters by $\cD\din$ and $\th$,
respectively, and the out-of-domain data and parameters by $\cD\dou$
and $\et$, we obtain the following form for these ``prior'' estimation
models:

\begin{equation}
\hat \th =
\arg\max_{\th}
\p{\th \| 
\arg\max_{\et}
\p{\et} \p{\cD\dou \| \et}}
\p{\cD\din \| \th}
\end{equation}

One would have a very difficult time rationalizing this optimization
problem by anything other than experimental performance.  Moreover,
these models are unusual in that they do not treat the in-domain data
and the out-of-domain data identically.  Intuitively, there is no
difference in the two sets of data; they simply come from different,
related distributions.  Yet, the prior-based models are highly
asymmetric with respect to the two data sets.  This also makes
generalization to more than one ``out of domain'' data set difficult.
Finally, as we will see, the model we propose in this paper, which
alleviates all of these problems, outperforms them experimentally.

A second generic approach to the domain adaptation problem is to build
an out of domain model and use its predictions as features for the in
domain data.  This has been successfully used in the context of named
entity tagging \cite{florian04edt}.  This approach is attractive
because it makes no assumptions about the underlying classifier; in
fact, multiple classifiers can be used.

\subsection{Our Framework} \label{sec:ourframework}

In this paper, we propose the following relationship between the
in-domain and the out-of-domain distributions.  We assume that instead
of two underlying distributions, there are actually \emph{three}
underlying distributions, which we will denote $q\dou$, $q\dgg$ and
$q\din$.  We then consider $p\dou$ to be a \emph{mixture} of $q\dou$
and $q\dgg$, and consider $p\din$ to be a mixture of $q\din$ and
$q\dgg$.  One can intuitively view the $q\dou$ distribution as a
distribution of data that is \emph{truly out-of-domain}, $q\din$ as a
distribution of data that is \emph{truly in-domain} and $q\dgg$ as a
distribution of data that is general to both domains.  Thus, knowing
$q\dgg$ and $q\din$ is sufficient to build a model of the in-domain
data.  The out-of-domain data can help us by providing more
information about $q\dgg$ than is available by just considering the
in-domain data.

For example, in part-of-speech tagging, the assignment of the tag
``determiner'' (DT) to the word ``the'' is likely to be a
\emph{general} decision, independent of domain.  However, in the Wall
Street Journal, ``monitor'' is almost always a verb (VB), but in
technical documentation it will most likely be a noun.  The $q\dgg$
distribution should account for the case of ``the/DT'', the $q\dou$
should account for ``monitor/VB'' and $q\din$ should account for
``monitor/NN.''

\section{Domain Adaptation in Maximum Entropy Models} \label{sec:megam}

The domain adaptation framework outlined in
Section~\ref{sec:ourframework} is completely general in that it can be
applied to any statistical learning model.  In this section we apply
it to log-linear conditional maximum entropy models and their linear
chain counterparts, since these models have proved quite effective in
many learning tasks.  We will first review the maximum entropy
framework, then will extend it to the domain adaptation problem;
finally we will discuss domain adaptation in linear chain maximum
entropy models.

\subsection{Maximum Entropy Models}

The maximum entropy framework seeks a conditional distribution $\p{y
\| x}$ that is closest (in the sense of KL divergence) to the uniform
distribution but also matches a set of training data $\D$ with respect
to feature function expectations \cite{dellapietra97inducing}.  By
introducing one Lagrange multiplier $\la_i$ for each feature function
$f_i$, this optimization problem results in a probability distribution
of the form:

\begin{equation} \label{eq:exponential}
\p{y \| x \; \vec\la} = \frac 1 {Z_{\vec\la,x}} \exp \left[\vec\la\T \vec f(x,y)\right]
\end{equation}

\begin{figure}
\center\mbox{
\psfig{figure=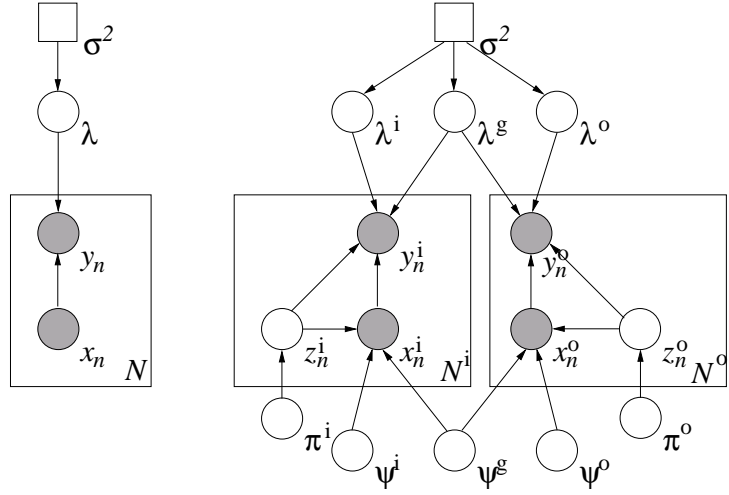,height=6.5cm}
}
\caption{(Left) the standard logistic regression model; (Right) the \megam.}
\label{fig:models}
\end{figure}

\noindent Here, $\vec u\T\vec v$ denotes the scalar product of two
vectors $\vec u$ and $\vec v$, given by: $\vec u\T\vec v = \sum_i u_i
v_i$.  The normalization constant in Eq~\eqref{eq:exponential},
$Z_{\vec\la,x}$, is obtained by summing the exponential over all
possible classes $y' \in \mathcal{Y}$.  This probability distribution
is also known as an \emph{exponential distribution} or a \emph{Gibbs
distribution}.  The learning (or optimization) problem is to find the
vector $\vec\la$ that maximizes the likelihood in
Eq~\eqref{eq:exponential}.  In practice, to prevent over-fitting, one
typically optimizes a penalized (log) likelihood, where an isotropic
Gaussian prior with mean $0$ and covariance matrix $\si^2 I$ is placed
over the parameters $\vec\la$ \cite{chen99gaussian}.  The graphical
model for the standard maximum entropy model is depicted on the left
of Figure~\ref{fig:models}.  In this figure, circular nodes correspond
to random variables and square nodes correspond to fixed variables.
Shaded nodes are observed in the training data and empty nodes are
hidden or unobserved.  Arrows denote conditional dependencies.

In general, the feature functions $f(x,y)$ may be arbitrary
real-valued functions; however, in this paper we will restrict our
attention to binary features.  In practice, this is not a harsh
restriction: many problems in the natural language domain naturally
employ only binary features (for real valued features, binning
techniques can be applied).  Additionally, for notational convenience,
we will assume that the features $f_i(x,y)$ can be written in product
form as $g_i(y)h_i(x)$ for arbitrary binary functions $g$ over outputs
and binary features $h$ over inputs.  The latter assumption means that
we can consider $x$ to be a binary vector where $x_i = h_i(x)$; in the
following this will simplify notation significantly (the extension to
the full case is straightforward, but messy, and is therefore not
considered in the remainder of this paper).  By considering $\vec x$
as a vector, we may move the class dependence to the parameters and
consider $\vec\la$ to be a matrix where $\la_{y,i}$ is the weight for
$h_i$ for class $y$.  We will write $\vec\la_y$ to refer to the column
vector of $\la$ corresponding to class $y$.  As $\vec x$ is also
considered a column vector, we write $\vec \la_y\T\vec x$ as shorthand
for the dot product between $\vec x$ and the weights for class $y$.
Under this modified notation, we may rewrite Eq~\eqref{eq:exponential}
as:

\begin{equation} \label{eq:maxent}
\p{y \| \vec x \; \vec\la} =
  \frac 1 {Z_{\vec\la,\vec x}} \exp \left[\vec\la_y\T \vec x\right]
\end{equation}

Combining this with a Gaussian prior on the weights, we obtain the
following form for the log posterior of a data set:

\begin{equation} \label{eq:maxent-post}
l = \log \p{\vec\la \| \cD, \si^s} =
- \frac 1 {2\si^2} \vec\la\T\vec\la +
\sum_{n=1}^N \left[
\vec\la_{y_n}\T\vec x_n -
\log \sum_{y' \in \cY}
\exp \left[ \vec\la_{y'}\T\vec x_n\right]
\right]
+ \textrm{const}
\end{equation}

The parameters $\vec\la$ can be estimated using any convex
optimization technique; in practice, limited memory BFGS
\cite{nash91bfgs,averick94evaluation} seems to be a good choice
\cite{malouf02opt,minka03logreg} and we will use this algorithm for
the experiments described in this paper.  In order to perform these
calculations, one must be able to compute the gradient of
Eq~\eqref{eq:maxent-post} with respect to $\vec \la$, which is
available in closed form.




\subsection{The Maximum Entropy Genre Adaptation Model} \label{subsec:megam}

Extending the maximum entropy model to account for both in-domain and
out-of-domain data in the framework described earlier requires the
addition of several extra model parameters.  In particular, for each
in-domain data point $(\vec x\din_n,y\din_n)$, we assume the existence
of a binary indicator variable $z\din_n$.  A value $z\din_n=1$
indicates that $(\vec x\din_n,y\din_n)$ is drawn from $q\din$ (the
truly in-domain distribution), while a value $z\din_n=0$ indicates
that it is drawn from $q\dgg$ (the general-domain distribution).
Similarly, for each out-of-domain data point $(\vec x\dou_n,
y\dou_n)$, we assume a binary indicator variable $z\dou_n$, where
$z\dou_n = 1$ means this data point is drawn from $q\dou$ (the truly
out-of-domain distribution) and a value of $0$ means that it is drawn
from $q\dgg$ (the general-domain distribution).  Of course, these
indicator variables are not observed in the data, so we must infer
their values automatically.

According to this model, the $z_n$s are binary random variables that
we assume are drawn from a Bernoulli distribution
with parameter $\pi\din$ (for in-domain) and $\pi\dou$ (for
out-of-domain).  Furthermore, we assume that there are three $\vec\la$
vectors, $\vec\la\din$, $\vec\la\dou$ and $\vec\la\dgg$ corresponding
to $q\din$, $q\dou$ and $q\dgg$, respectively.  For instance, if
$z_n=1$, then we assume that $\bar x_n$ should be classified using
$\vec\la\din$.  Finally, we model the binary vectors $\vec x\din_n$s
(respectively $\vec x\dou_n$s) as being drawn independently from
Bernoulli distributions parameterized by $\vec \ps\din$ and $\vec
\ps\dgg$ (respectively, $\vec \ps\dou$ and $\vec \ps\dgg$).  Again,
when $z_n = 1$, we assume that $\vec x_n$ is drawn according to $\vec
\ps\din$.  This corresponds to a na\"ive Bayes assumption over the
generative probabilities of the $\vec x_n$ vectors.  Finally, we place
a common Beta prior over the na\"ive Bayes
parameters, $\vec\ps$.  Allowing $\nu$ to range over $\{
\textsf{i},\textsf{o},\textsf{g} \}$, the full hierarchical model is:

\begin{small}
\begin{hierarchical2col}
\vec\ps\dnu_f   & a, b     & \Bet(a,b)  &
\vec\la\dnu     & \si^2    & \Nor(0, \si^2 I) \\
z\din_n     & \pi\din  & \Ber(\pi\din) &
z\dou_n    & \pi\dou & \Ber(\pi\dou) \\
x\din_{nf}  & z\din_n , \vec\ps\din_f , \vec\ps\dgg_f & \Ber(\vec\ps^{z\din_n}_f) &
x\dou_{nf} & z\dou_n, \vec\ps\dou_f, \vec\ps\dgg_f & \Ber(\vec\ps^{z\dou_n}_f) \\
y\din_n  & x\din_n , z\din_n , \vec\la\din , \vec\la\dgg & \textit{Gibbs}(x\din_n, \vec\la^{z\din_n}) &
y\dou_n & x\dou_n, z\dou_n, \vec\la\dou, \vec\la\dgg & \textit{Gibbs}(x\dou_n,\vec\la^{z\dou_n}) 
\label{eq:hier-megam}
\end{hierarchical2col}
\end{small}

We term this model the ``Maximum Entropy Genre Adaptation Model'' (the
\megam).  The corresponding graphical model is shown on the right in
Figure~\ref{fig:models}.  The generative story for an in-domain data
point $\vec x\din$ is as follows:

\begin{enumerate}
\item Select whether $\vec x\din$
will be truly in-domain or general-domain and indicate this by $z\din
\in \{ i, g \}$.  Choose $z\din = i$ with probability $\pi\din$ and
$z\din = g$ with probability $1-\pi\din$.  
\item For each component $f$ of $\vec x\din$, choose $x\din_f$ to be $1$ with probability
$\vec\ps^{z\din}_f$ and $0$ with probability $1-\vec\ps^{z\din}_f$.
\item Choose a class $y$ according to Eq~\eqref{eq:maxent} using
the parameter vector $\vec\la^{z\din}$.  
\end{enumerate}

The story for out-of-domain data points is identical, but uses the
truly out-of-domain and general-domain parameters, rather than the
truly in-domain parameters and general-domain parameters.

\subsection{Linear Chain Models}

The straightforward extension of the maximum entropy classification
model to the maximum entropy Markov model (MEMM) \cite{mccallum00memm}
is obtained by assuming that the targets $y_n$ are sequences of
labels.  The canonical example for this model is part of speech
tagging: each word in a sequence is assigned a part of speech tag.  By
introducing a first order Markov assumption on the tag sequence, one
obtains a linear chain model that can be viewed as the discriminative
counterpart to the standard (generative) hidden Markov model.  The
parameters of these models can be estimated again using limited memory
BFGS.  The extension of the \megam\ to the linear chain framework is
similarly straightforward, under the assumption that each label (part
of speech tag) has its own indicator variable $z$ (versus a global
indicator variable $z$ for the entire tag sequence).

The techniques described herein may also be applied to the conditional
random field framework of \citeA{lafferty01crf}, which fixes a bias
problem of the MEMM by performing global normalization rather than
per-state normalization.  There is, however, a subtle difficulty in a
direct application to CRFs.  Specifically, one would need to decide if
a single $z$ variable would be assigned to an entire sentence, or to
each word individually.  In the MEMM case, it is most natural to have
one $z$ per word.  However, to do so in a CRF would be computationally
more expensive.  In the remainder, we continue to use the MEMM model
for efficiency purposes.

\section{Conditional Expectation Maximization} \label{sec:cem}

Inference in the \megam\ is slightly more complex than in standard
maximum entropy models.  However, inference can be solved efficiently
using conditional expectation maximization (CEM), a variant of the
standard expectation maximization (EM) algorithm \cite{dempster77em},
due to \citeA{jebara98cem}.  At a high level, EM is useful for
computing in \emph{generative} models with hidden variables, while CEM
is useful for computing in \emph{discriminative} models with hidden
variables; the \megam\ belongs to the latter family, so CEM is the
appropriate choice.

The standard EM family of algorithms maximizes a \emph{joint}
likelihood over data.  In particular, if $(x_n,y_n)_{n=1}^N$ are data
and $z$ is a (discrete) hidden variable, the M-step of EM proceeds by
maximizing the bound given in Eq~\eqref{eq:em-joint}

\begin{equation} \label{eq:em-joint}
\log \p{x, y \| \Th} = \log \sum_z \p{z,x,y \| \Th} =
\log \Ep_{z \sim \p{\cdot \| x; \Th}} \p{x,y \| z; \Th}
\end{equation}

In Eq~\eqref{eq:em-joint}, $\Ep_z$ denotes an expectation.  One may now
apply Jensen's inequality to this equation, which states that
$f(\Ep\{x\}) \leq \Ep\{f(x)\}$ whenever $f$ is convex.  Taking
$f=\log$, we are able to decompose the $\log$ of an expectation into
the expectation of a $\log$.  This typically separates terms and makes
taking derivatives and solving the resolution optimization problem
tractable.  Unfortunately, EM cannot be directly applied to
conditional models (such as the \megam) of the form in
Eq~\eqref{eq:conditional} because such models result in an M-step that
requires the maximization of an equation of the form given in
Eq~\eqref{eq:cem-lik}.

\begin{align}
\log \p{y \| x; \Th} &= \log \sum_z \p{z, y \| x; \Th}
= \log \Ep_{z \sim \p{\cdot \| x,\Th}} \p{y \| x, z; \Th} \label{eq:conditional}\\
l &= 
  \log \sum_z \p{z,x,y \| \Th} -
  \log \sum_z \p{z,x \| \Th} \label{eq:cem-lik}
\end{align}

Jensen's inequality can be applied to the first term in
Eq~\eqref{eq:cem-lik}, which can be maximized readily as in standard
EM.  However, applying Jensen's inequality to the second term would
lead to an \emph{upper bound} on the likelihood, since that term
appears negated.

The conditional EM solution \cite{jebara98cem} is to bound the
\emph{change} in log-likelihood between iterations, rather than the
log-likelihood itself.  The change in log-likelihood can be written as
in Eq~\eqref{eq:de-l-c0}, where $\Th^t$ denotes the parameters at
iteration $t$.

\begin{equation} \label{eq:de-l-c0}
\De l^c = \log \p{y \| x ; \Th^t} - \log \p{y \| x ; \Th^{t-1}}
\end{equation}

By rewriting the conditional distribution $\p{y \| x}$ as $\p{x,y}$
divided by $\p{x}$, we can express $\De l^c$ as the log of the joint
distribution difference minus the log of the marginal distribution.
Here, we can apply Jensen's inequality to the first term (the joint
difference), but not to the second (because it appears negated).
Fortunately, Jensen's is not the only bound we can employ.  The
standard variational upper bound of the logarithm function is: $\log x
\leq x-1$; this leads to a lower bound of the negation, which is
exactly what is desired.  This bound is attractive for other reasons:
(1) it is tangent to the logarithm; (2) it is tight; (3) it makes
contact at the current operating point (according to the maximization
at the previous time step); (4) it is a simply linear function; and
(5) in the terminology of the calculus of variations, it is the
variational dual to the logarithm; see \cite{smith98variational}.

Applying Jensen's inequality to the first term in
Eq~\eqref{eq:de-l-c0} and the variational dual to the second term, we
obtain that the change of log-likelihood in moving from model
parameters $\Th^{t-1}$ at time $t-1$ to $\Th^t$ at time $t$ (which we
shall denote $Q^t$) is bounded by $\De l \geq Q^t$, where $Q^t$ is
defined by Eq~\eqref{eq:Q}, where $h = \Ep\{z \| x; \Th\}$ when $z=1$ and
$1-\Ep\{z \| x; \Th\}$ when $z=0$, with expectations taken with respect to the
parameters from the previous iteration.

\begin{equation} \label{eq:Q}
Q^t =
\sum_{z \in \cZ}
    h_z
    \log \frac {\p{z, x, y \| \Th^t}} {\p{z, x, y \| \Th^{t-1}}}
  - \frac {\sum_{z} \p{z,x \| \Th^t}} {\sum_{z} \p{z,x \| \Th^{t-1}}}
  + 1
\end{equation}

By applying the two bounds (Jensen's inequality and the variational
bound), we have removed all ``sums of logs,'' which are hard to deal
with analytically.  The full derivation is given in Appendix A.  The
remaining expression is a lower bound on the change in likelihood, and
maximization of it will result in maximization of the likelihood.

As in the MAP variant of standard EM, there is no change to the E-step
when priors are placed on the parameters.  The assumption in standard
EM is that we wish to maximize $\p{\Th \| \vec x, \vec y} \varpropto
\p{\Th} \p{\vec y \| \Th, \vec x}$ where the prior probability of
$\Th$ is ignored, leaving just the likelihood term of the parameters
given the data.  In MAP estimation, we do not make this assumption and
instead use a true prior $\p{\Th}$.  In doing so, we need only to add
a factor of $\log \p{\Th}$ to the definition of $Q^t$ in
Eq~\eqref{eq:Q}.

It is important to note that although we do make use of a full joint
distribution $\p{x,y,z}$, the \emph{objective function} of our model
is \emph{conditional.}  The joint distribution is only used in the
process of creating the bound: the overall optimization is to maximize
the conditional likelihood of the labels \emph{given} the input.  In
particular, the bound using the full joint likelihood holds for any
parameters of the marginal.

\section{Parameter Estimation for the \megam} \label{sec:estimation}

As made explicit in Eq~\eqref{eq:Q}, the relevant distributions for
performing CEM are the full joint distributions over the input
variables $x$, the output variables $y$, and the hidden variables $z$.
Additionally, we require the marginal distribution over the $x$
variables and the $z$ variables.  Finally, we need to compute
expectations over the $z$ variables.  We will derive the expectation
step in this section and present the final solution for the
maximization step for each class of variables.  The derivation of the
equations for the maximization is given in Appendix B.

The $Q$ bound on complete conditional likelihood for the \megam is
given below:

\begin{align} 
Q^t &=
  \sum_{n=1}^{N\din} \left[\sum_{z\din_n} 
    h\din_n
    \hspace{0.7mm}
    \log \frac {\p{z\din_n\hspace{0.7mm}, \vec x\din_n\hspace{0.7mm}, y\din_n\hspace{0.7mm}}} {\pprime{z\din_n\hspace{0.7mm}, \vec x\din_n\hspace{0.7mm}, y\din_n\hspace{0.7mm}}}
\hspace{0.2mm}
  - \frac {\sum_{z\din_n} \p{z\din_n\hspace{0.9mm},\vec x\din_n\hspace{0.9mm}}} {\sum_{z\din_n} \pprime{z\din_n\hspace{0.9mm},\vec x\din_n\hspace{0.9mm}}}
  \hspace{0.4mm}+ 1 \right] \nonumber\\
&+
  \sum_{n=1}^{N\dou} \hspace{0.3mm} \left[\sum_{z\dou_n} 
    h\dou_n
    \log \frac {\p{z\dou_n, \vec x\dou_n, y\dou_n}} {\pprime{z\dou_n, \vec x\dou_n, y\dou_n}}
  - \frac {\sum_{z\dou_n} \p{z\dou_n,\vec x\dou_n}} {\sum_{z\dou_n} \pprime{z\dou_n,\vec x\dou_n}}
  + 1 \right] \label{eq:Qmegam}
\end{align}

In this equation, $\pprime{}$ is the probability distribution at the
previous iteration.  The first term in Eq~\eqref{eq:Qmegam} is the
bound for the in-domain data, while the second term is the bound for
the out-of-domain data.  In all the optimizations described in this
section, there are nearly identical terms for the in-domain parameters
and the out-of-domain parameters.  For brevity, we will only
explicitly write the equations for the in-domain parameters; the
corresponding out-of-domain equations can be easily derived from
these.  Moreover, to reduce notational overload, we will elide the
superscripts denoting in-domain and out-of-domain when obvious from
context.  For notational brevity, we will use the notation depicted in
Table~\ref{tab:notation}.

\begin{table}
\begin{equation} \label{eq:notation}
\begin{array}{rclrcl}
j_{n,z_n}^{t-1} &=& \log \p{x_n, y_n, z_n \| \Th^{t-1}} &
\ps_{n,z_n}     &=& \prod_{f=1}^F \hspace{1mm}\left(\ps^{z_n}_f\right)^{x_{nf}}
      \left(1-\ps^{z_n}_f\right)^{1-x_{nf}} \nonumber\\
m_{n}^{t-1}     &=& \left[ {\sum_{z_n} \p{x_n,z_n \|
                   \Th^{t-1}}} \right]^{-1} &
\ps_{n,z_n,-f'}     &=& \prod_{f \ne f'} \left(\ps^{z_n}_{f}\right)^{x_{nf}}
      \left(1-\ps^{z_n}_{f}\right)^{1-x_{nf}} \nonumber
\end{array}
\end{equation}
\caption{Notation used for \megam\ equations.}
\label{tab:notation}
\end{table}

\subsection{Expectation Step}

The E-step is concerned with calculating $h_n$ given current model
parameters.  Since $z_n \in \{ 0, 1\}$, we easily find $h_n = \p{z_n =
1 | \Th}$, which can be calculated as follows:

\begin{eqnarray}
&&\hspace{-20mm}\p{z_n = z \| \vec x_n, y_n, \vec\ps, \vec\la, \pi}  \nonumber\\
&=& \frac {\p{z_n=z \| \pi} \p{\vec x_n \| \vec\ps, z_n=z} \p{y_n \| \vec\la, z_n=z}}
         {\sum_z ~\p{z_n=z \| \pi} \p{\vec x_n \| \vec\ps, z_n=z} \p{y_n \| \vec\la, z_n=z}}
  \nonumber\\
&\varpropto&
  \pi^z (1-\pi)^{1-z} \ps_{n,z}
  \frac 1 {Z_{\vec x_n,\vec\la^z}} \exp\left[\vec\la^z_{y_n} \T \vec x_n\right] \label{eq:estep}
\end{eqnarray}

\noindent
Here, $Z$ is the partition function from before.  This can be easily
calculated for $z \in \{ 0, 1 \}$ and the expectation can be found by
dividing the value for $z=1$ by the sum over both.

\subsection{M-Step for $\pi$}\label{sec:msteppi}

As shown in Appendix B.1, we can directly compute the value of $\pi$
by solving a simple quadratic equation.  We can compute $\pi$ as $-a +
\sqrt{a^2 - b}$, where:

\begin{eqnarray*}
a &=& 
\frac {1 - \sum_{n=1}^N 
        \(2 h_n - m_n^{t-1} \left(\ps_{n,0} - \ps_{n,1}\right)\)}
      {2 \sum_{n=1}^N m_n^{t-1} \left(\ps_{n,0} - \ps_{n,1}\right)}\\
b &=& 
-\frac {\sum_{n=1}^N h_n} {\sum_{n=1}^N m_n^{t-1} \left(\ps_{n,0} - \ps_{n,1}\right)}
\end{eqnarray*}

\subsection{M-Step for $\vec\la$} \label{sec:mstepphi}

Viewing $Q^t$ as a function of $\vec\la$, it is easy to see that
optimization for this variable is convex.  An analytical solution is
not available, but the gradient of $Q^t$ with respect to $\vec\la\din$
can be seen to be identical to the gradient of the standard maximum
entropy posterior, Eq~\eqref{eq:maxent-post}, but where each data
point is \emph{weighted} according to its posterior probability,
$(1-h_n)$.  We may thus use identical optimization techniques for
computing optimal $\vec\la$ variables as for standard maximum entropy
models; the only difference is that the data points are now weighted.
A similar story holds for $\vec\la\dou$.  In the case of
$\vec\la\dgg$, we obtain the standard maximum entropy gradient,
computed over all $N\din + N\dou$ data points, where each $x\din_n$ is
weighted by $h_n$ and each $x\dou_n$ is weighted by $h\dou_n$.  This
is shown in Appendix B.2.

\subsection{M-Step for $\vec\ps$} \label{sec:msteppsi}

Like the case for $\vec\la$, we cannot obtain an analytical solution
for finding the $\vec\ps$ that maximizes $Q^t$.  However, we can
compute simple derivatives for $Q^t$ with respect to a single
component $\vec\ps_f$ which can be maximized analytically.  As
shown in Appendix B.3, we can compute $\ps\din_f$ as $-a +
\sqrt{a^2 - b}$, where:

\begin{eqnarray*}
a &=& -\frac {\sum_{n=1}^N \Big(1-h_n + j_{n,0}(1-\pi) \ps_{n,0,-f}\Big)}
       {2 \sum_{n=1}^N j_{n,0} (1-\pi) \ps_{n,0,-f}} \\
b &=& \frac {1 + \sum_{n=1}^N \(1-h_n\) x_{nf}}
       {\sum_{n=1}^N j_{n,0} (1-\pi) \ps_{n,0,-f}}
\end{eqnarray*}

The case for $\vec\ps\dou$ is identical.  For $\vec\ps\dgg$, the only
difference is that we must replace each sum to over the data points
with two sums, one for each of the in-domain and out-of-domain points;
and, as before, the $1-h_n$s must be replaced with $h_n$; this is made
explicit in the Appendix.  Thus, to optimize the $\vec\ps$ variables,
we simply iterate through and optimize each component analytically,
as given above, until convergence.

\subsection{Training Algorithm}

\begin{figure}[t]
\framebox{\begin{minipage}[t]{\textwidth}
\begin{algorithmic}
\STATE {\bf Algorithm} \textsc{MegaCEM}
\STATE Initialize $\ps\dnu_f = 0.5$, $\vec\la\dnu_f = 0$, $\pi\dnu
= 0.5$ for all $\nu \in \{ g,i,o \}$ and all $f$.
\WHILE{parameters haven't converged or iterations remain}
\STATE{}
\STATE{\textsf{\{- Expectation Step -\}}}
\FOR{$n = 1..N\din$}
\STATE Compute the in-domain marginal probabilities, $m\din_n$
\STATE Compute the in-domain expectations, $h\din_n$, by Eq~\eqref{eq:estep}
\ENDFOR
\FOR{$n = 1..N\dou$}
\STATE Compute the out-of-domain marginal probabilities, $m\dou_n$
\STATE Compute the out-of-domain expectations,  $h\dou_n$ by Eq~\eqref{eq:estep}
\ENDFOR
\STATE{}
\STATE{\textsf{\{- Maximization Step -\}}}
\STATE Analytically update $\pi\din$ and $\pi\dou$ according to the
equations shown in Section~\ref{sec:msteppi}
\STATE Optimize $\vec\la\din$, $\vec\la\dou$ and $\vec\la\dgg$ using BFGS
\WHILE{Iterations remain and/or $\vec\ps$ haven't converged}
\STATE Update $\vec\ps$s according to derivation in Section~\ref{sec:msteppsi}
\ENDWHILE
\STATE{}
\ENDWHILE
\STATE{{\bf return} $\vec\la, \vec\ps, \pi$}
\end{algorithmic}
\end{minipage}}
\caption{The full training algorithm for the \megam.}
\label{fig:training-algorithm}
\end{figure}

The full training algorithm is depicted in
Figure~\ref{fig:training-algorithm}.  Convergence properties of the
CEM algorithm ensure that this will converge to a (local) maximum in
the posterior space.  If local optima become a problem in practice,
one can alternatively use a stochastic optimization algorithm, in
which a temperature is applied enabling the optimization to jump out
of local optima early on.  However, we do not explore this idea
further in this work.  In the context of our application, this
extension was not required.

\subsection{CEM Convergence}

\begin{figure}[!t]
\center\mbox{
\psfig{figure=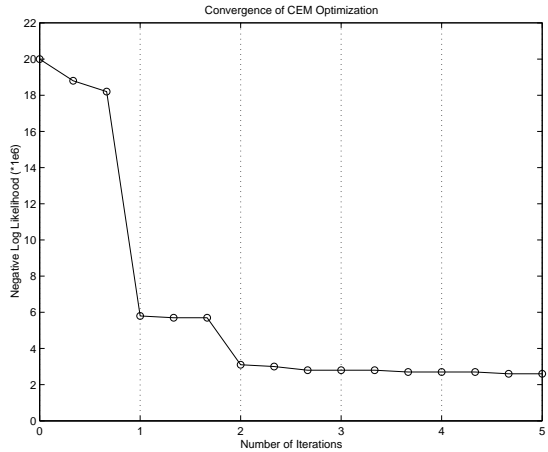,height=6cm}
}
\caption{Convergence of training algorithm.}
\label{fig:cll}
\end{figure}

One immediate question about the conditional EM model we have
described is how many EM iterations are required for the model to
converge.  In our experiments, $5$ iterations of CEM is more than
sufficient, and often only $2$ or $3$ are necessary.  To make this
more clear, in Figure~\ref{fig:cll}, we have plotted the negative
complete log likelihood of the model on the first data set, described
below in Section~\ref{datasets}.  There are three separate
maximizations in the full training algorithm (see
Figure~\ref{fig:training-algorithm}); the first involves updating the
$\pi$ variables, the second involves optimizing the $\vec\la$
variables and the third involves optimizing the $\vec\ps$ variables.
We compute the likelihood after each of these steps.

Running a total $5$ CEM iterations is still relatively efficient in
our model.  The dominating expense is in the weighted maximum entropy
optimization, which, at $5$ CEM iterations, must be computed $15$
times (each iteration requires the optimization of each of the three
sets of $\vec\la$ variables).  At worst this will take $15$ times the
amount of time to train a model on the complete data set (the union of
the in-domain and out-of-domain data), but in practice we can resume
each optimization at the ending point of the previous iteration, which
causes the subsequent optimizations to take much less time.

\subsection{Prediction}

Once training has supplied us with model parameters, the subsequent
task is to apply these parameters to unseen data to obtain class
predictions.  We assume all this test data is ``in-domain'' (i.e., is
drawn either from $Q\din$ or $Q\dgg$ in the notation of the
introduction), and obtain a decision rule of the form given in
Eq~\eqref{eq:predict} for a new test point $\vec x$.

\begin{align} 
\hat y 
&= \arg\max_{y \in \cY} \p{y \| x; \Th} \nonumber\\
&= \arg\max_{y \in \cY} \sum_z \p{z \| x; \Th} \p{y \| x,z; \Th} \nonumber\\
&= \arg\max_{y \in \cY} \sum_z \p{z \| \Th} \p{x \| z; \Th} \p{y \| x,z; \Th} \nonumber\\
&= \arg\max_{y \in \cY} ~
\pi \left[\prod_{f=1}^F {\(\ps\dgg_f\)^{x_f}} {\(1-\ps\dgg_f\)^{1-x_f}}
    \hspace{0.5mm} \right]
    \frac {\exp\left[\vec\la_y\dgg \T \vec x\right]} {Z_{\vec x, \vec\la\dgg}}
\nonumber\\
&
\hspace{6.0mm}
+(1-\pi) \left[\prod_{f=1}^F {\(\ps\din_f\hspace{0.7mm}\)^{x_f}}
  {\(1-\ps\din_f \hspace{0.8mm}\)^{1-x_f}} \hspace{.5mm}\right]
    \frac {\exp\left[\vec\la_y\din \T \vec x\right]} {Z_{\vec x, \vec\la\din}} 
\label{eq:predict}
\end{align}

Thus, the decision rule is to simply select the class which has
highest probability according to the maximum entropy classifiers,
weighted linearly by the marginal probabilities of the new data point
being drawn from $Q\din$ versus $Q\dgg$.  In this sense, our model can be
seen as linearly interpolating an in-domain model and a general-domain
model, but where the interpolation parameter is \emph{input specific}.

\section{Experimental Results} \label{sec:experiments}

In this section, we describe the result of applying the \megam\ to
several datasets with varying degrees of divergence between the
in-domain and out-of-domain data.  However, before describing the data
and results, we will discuss the systems against which we compare.

\subsection{Baseline Systems}

Though there has been little literature on this problem and thus few
real systems against which to compare, there are several obvious
baselines, which we describe in this section.

\paragraph{\system{OnlyI}:} This model is obtained simply by training
a standard maximum entropy model on the \emph{in-domain} data.  This
completely ignores the out-of-domain data and serves as a baseline
case for when such data is unavailable.

\paragraph{\system{OnlyO}:} This model is obtained by training a
standard maximum entropy model on the \emph{out-of-domain} data,
completely ignoring the in-domain data.  This serves as a baseline for
expected performance without annotating any new data.  It also gives a
sense of how close the out-of-domain distribution is to the in-domain
distribution.

\paragraph{\system{LinI}:} This model is obtained by linearly
interpolating the \system{OnlyI} and \system{OnlyO} systems.  The
interpolation parameter is estimated on held-out (development)
in-domain data.  This means that, in practice, extra in-domain data
would need to be annotated in order to create a development set;
alternatively, cross-validation could be used.

\paragraph{\system{Mix}:} This model is obtained by training a maximum
entropy model on the union of the out-of-domain and in-domain data
sets.

\paragraph{\system{MixW}:} This model is also obtained by training a
maximum entropy model on the union of the out-of-domain and in-domain
data sets, but where the out-of-domain data is \emph{down-weighted} so
that is effectively equinumerous with the in-domain data.

\paragraph{\system{Feats}:} This model uses the out-of-domain data to
build one classifier and then uses this classifier's predictions as
features for the in-domain data, as described by \citeA{florian04edt}.

\paragraph{\system{Prior}:} This is the adaptation model described in
Section~\ref{prior-work}, where the out-of-domain data is used to
estimate a prior for the in-domain classifier.  In the case of the
maximum entropy models we consider here, the weights learned from the
out-of-domain data are used as the \emph{mean} of the Gaussian prior
distribution placed over the weights in the training of the in-domain
data, as is described by \citeA{chelba04adapt}.

In all cases, we tune model hyperparameters using performance on
development data.  This development data is taken to be a random
$20\%$ of the training data in all cases.  Once appropriate
hyperparameters are found, the $20\%$ is folded back in to the
training set.

\subsection{Data Sets} \label{datasets}

We evaluate our models on three different problems.  The first two
problems come from the Automatic Content Extraction (ACE) data task.
This data was selected because the ACE program specifically looks at
data in different domains.  The third problem is the same as that
tackled by \citeA{chelba04adapt}, which required them to annotate data
themselves.

\subsubsection{Mention Type Classification} \label{sec:type}

The first problem, {\bf Mention Type}, is a subcomponent of the entity
mention detection task (an extension of the named entity tagging task,
wherein pronouns and nominals are marked, in addition to simple
names).  We assume that the extents of the mentions are marked and we
simply need to identify their type, one of: Person, Geo-political
Entity, Organization, Location, Weapon or Vehicle.  As the
out-of-domain data, we use the newswire and broadcast news portions of
the ACE 2005 training data; as the in-domain data, we use the Fisher
conversations data.  An example out-of-domain sentence is:

\begin{quote}
Once again, a prime battleground will be the constitutional
allocation of power -- between the federal
\mention{government}{gpe}{nom} and the \mention{states}{gpe}{nom}, and
between \mention{Congress}{org}{nam} and federal regulatory
\mention{agencies}{org}{bar} .
\end{quote}

\noindent An example in-domain sentence is:

\begin{quote}
\mention{my}{per}{pro} \mention{wife}{per}{nom} if
\mention{I}{per}{pro} had not been transported across the
\mention{continent}{gpe}{nom} from \mention{where}{loc}{whq}
\mention{I}{per}{pro} was born and and
\end{quote}

We use $23k$ out-of-domain examples (each mention corresponds to one
example), $1k$ in-domain examples and $456$ test examples.  Accuracy
is computed as $0/1$ loss.  We use the standard feature functions
employed in named entity models, include lexical items, stems,
prefixes and suffixes, capitalization patterns, part-of-speech tags,
and membership information on gazetteers of locations, businesses and
people.  The accuracies reported are the result of running ten fold
cross-validation.

\subsubsection{Mention Tagging} \label{sec:tagging}

The second problem, {\bf Mention Tagging} is the precursor to the {\bf
Mention Type} task, in which we attempt to tag entity mentions in raw
text.  We use the standard Begin/In/Out encoding and use a maximum
entropy Markov model to perform the tagging \cite{mccallum00memm}.  As
the out-of-domain data, we use again the newswire and broadcast news
data; as the in-domain data, we use broadcast news data that has been
transcribed by automatic speech recognition.  The in-domain data lacks
capitalization, punctuation, etc., and also contains transcription
errors (speech recognition word error rate is approximately $15\%$).
For the tagging task, we have $112k$ out-of-domain examples (in the
context of tagging, an example is a single word), but now $5k$
in-domain examples and $11k$ test examples.  Accuracy is F-measure
across the segmentation.  We use the same features as in the mention
type identification task.  The scores reported are after ten fold
cross-validation.

\subsubsection{Recapitalization}

The final problem, {\bf Recap}, is the task of recapitalizing text.
Following \citeA{chelba04adapt}, we again use a maximum entropy Markov
model, where the possible tags are: Lowercase, Capitalized, All Upper
Case, Punctuation or Mixed case.  The out-of-domain data in this task
comes from the Wall Street Journal, and two separate in-domain data
sets come from broadcast news text from CNN/NPR and ABC Primetime,
respectively.  We use $3.5m$ out-of-domain examples (one example is
one word).  For the CNN/NPR data, we use $146k$ in-domain training
examples and $73k$ test examples; for the ABC Primetime data, we use
$33k$ in-domain training examples and $8k$ test examples.  We use
identical features to \citeA{chelba04adapt}.  In order to maintain
comparability to the results described by \citeA{chelba04adapt}, we do
not perform cross-validation for these experiments: we use the same
train/test split as described in their paper.

\subsection{Feature Selection}

While the maximum entropy models used for the classification are adept
at dealing with many irrelevant and/or redundant features, the na\"ive
Bayes generative model, which we use to model the distribution of the
input variables, can overfit on such features.  This turned out not to
be a problem for the {\bf Mention Type} and {\bf Mention Tagging}
problems, but for the {\bf Recap} problems, it caused some errors.  To
alleviate this problem, for the {\bf Recap} problem \emph{only}, we
applied a feature selection algorithm just to the features used for
the na\"ive Bayes model (the entire feature set was used for the
maximum entropy model).  Specifically, we took the $10k$ top features
according to the information gain criteria to predict ``in-domain''
versus ``out-of-domain'' (as opposed to feature selection for class
label); \citeA{forman03selection} provides an overview of different
selection techniques.\footnote{The value of $10k$ was selected
arbitrarily after an initial run of the model on development data; it
was not tuned to optimize either development or test performance.}

\subsection{Results}

\begin{table*}[!t]
\center
\begin{tabular}{|ll||c|c||c|c||c|}
\hline
&               & {\bf Mention} & {\bf Mention} & {\bf Recap} & {\bf Recap} &\\
&               & {\bf Type}    & {\bf Tagging} & {\bf ABC}   & {\bf CNN}   & {\bf Average} \\
\hline
\multicolumn{2}{|l||}{$|\D\dou|$} & $23k$ & $112k$ & $3.5m$ & $3.5m$ & -\\
\multicolumn{2}{|l||}{$|\D\din|$} &  $1k$ &   $5k$ &   $8k$ &  $73k$ & -\\
\hline
\multicolumn{2}{|l||}{\bf Accuracy}   & & & & & \\
&\system{OnlyO} & 57.6          & 78.3          & 95.5        & 94.6        & 81.5 \\
&\system{OnlyI} & 81.2          & 83.5          & 97.4        & 94.7        & 89.2 \\
&\system{LinI}  & 81.5          & 83.8          & 97.7        & 94.9        & 89.5 \\
&\system{Mix}   & 84.9          & 80.9          & 96.4        & 95.0        & 89.3 \\
&\system{MixW}  & 81.3          & 81.0          & 97.6        & 93.5        & 88.8 \\
&\system{Feats} & 87.8          & 84.2          & 97.8        & 96.1        & 91.5 \\
&\system{Prior} & 87.9          & 85.1          & 97.9        & 95.9        & 91.7 \\
&\system{MegaM} & 92.1          & 88.2          & 98.1        & 96.8        & 93.9 \\
\hline
\multicolumn{2}{|l||}{\bf \% Reduction} & & & & & \\
&\system{Mix}   & 47.7          & 38.2          & 52.8        & 36.0        & 43.0 \\
&\system{Prior} & 34.7          & 20.8          & 19.0        & 22.0        & 26.5 \\
\hline
\end{tabular}
\caption{Experimental results; The first set of rows show the sizes of
  the in-domain and out-of-domain training data sets.  The second set
  of rows (Accuracy) show the performance of the various models on
  each of the four tasks.  The last two rows (\% Reduction) show the
  percentage reduction in error rate by using the \megam\ over the
  baseline model ({\bf\textsc{Mix}}) and the best alternative method
  ({\bf\textsc{Prior}}).}
\label{fig:results}
\end{table*}


Our results are shown in Table~\ref{fig:results}, where we can see
that training only on in-domain data always outperforms training only
on out-of-domain data.  The linearly interpolated model does not
improve on the base models significantly.  Placing all the data in one
bag helps, and there is no clear advantage to re-weighting the out
domain data.  The \system{Prior} model and the \system{Feats} model
perform roughly comparably, with the \system{Prior} model edging out
by a small margin.\footnote{Our numbers for the result of the
\system{Prior} model on the data from \citeA{chelba04adapt} differ
slightly from those reported in their paper.  There are two potential
reasons for this.  First, most of their numbers are reported based on
using all $20m$ examples; we consider only the $3.5m$ example case.
Second, there are likely subtle differences in the training algorithms
used.  Nevertheless, on the whole, our relative improvements agree
with those in their paper.}  Our model outperforms both the
\system{Prior} model and the \system{Feats} model.

We applied McNemar's test \cite[section
14.5]{gibbons03statistical} to gage statistical significance of these
results, comparing the results of the \system{Prior} model with our
own \megam\ (for the mention tagging experiment, we compute McNemar's
test on simple Hamming accuracy rather than F-score; this is
suboptimal, but we do not know how to compute statistical significance
for the F-score).  For the mention type task, the difference is
statistical significant at the $p \leq 0.03$ level; for the mention
tagging task, $p \leq 0.001$; for the recapitalization tasks, the
difference on the ABC data is significant only at the $p \leq 0.06$
level, while for the CNN/NPR data it is significant at the $p \leq
0.004$ level.

In the mention type task, we have improved a baseline model trained
only on in-domain data from an accuracy of $81.2\%$ up to $92.1\%$, a
relative improvement of $13.4\%$.  For mention tagging, we improve
from $83.5\%$ F-measure up to $88.2\%$, a relative improvement of
$5.6\%$.  In the ABC recapitalization task (for which much in-domain
data is available), we increase performance from $95.5\%$ to $98.1\%$,
a relative improvement of $2.9\%$.  In the CNN/NPR recapitalization
task (with very little in-domain data), we increase performance from
$94.6\%$ to $96.8\%$, a relative improvement of $2.3\%$.

\subsection{Learning Curves}

\begin{figure*}[!t]
\center\mbox{
\hspace{-5mm}
\begin{tabular}{lr}
\psfig{figure=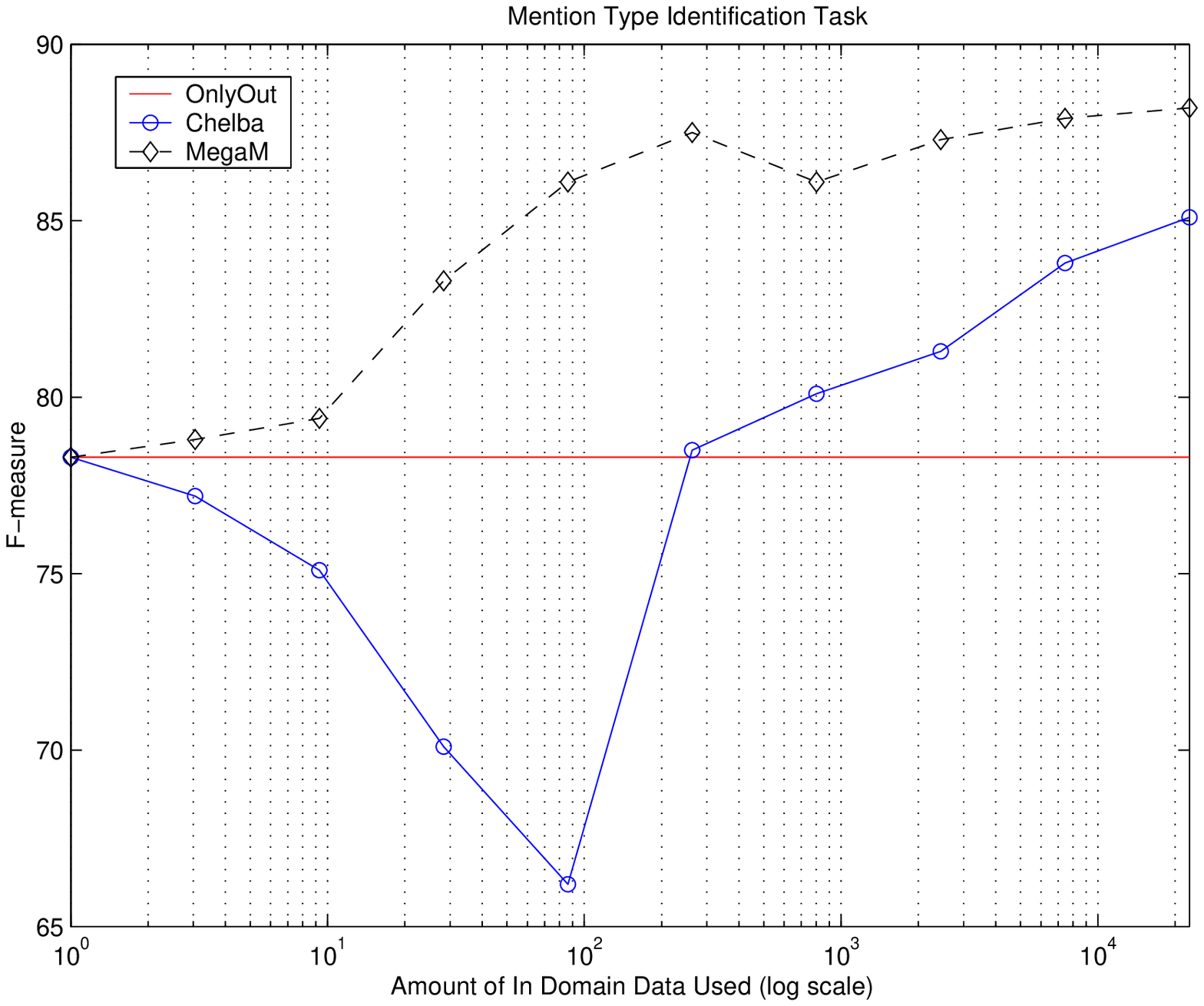,width=7.5cm} &
\psfig{figure=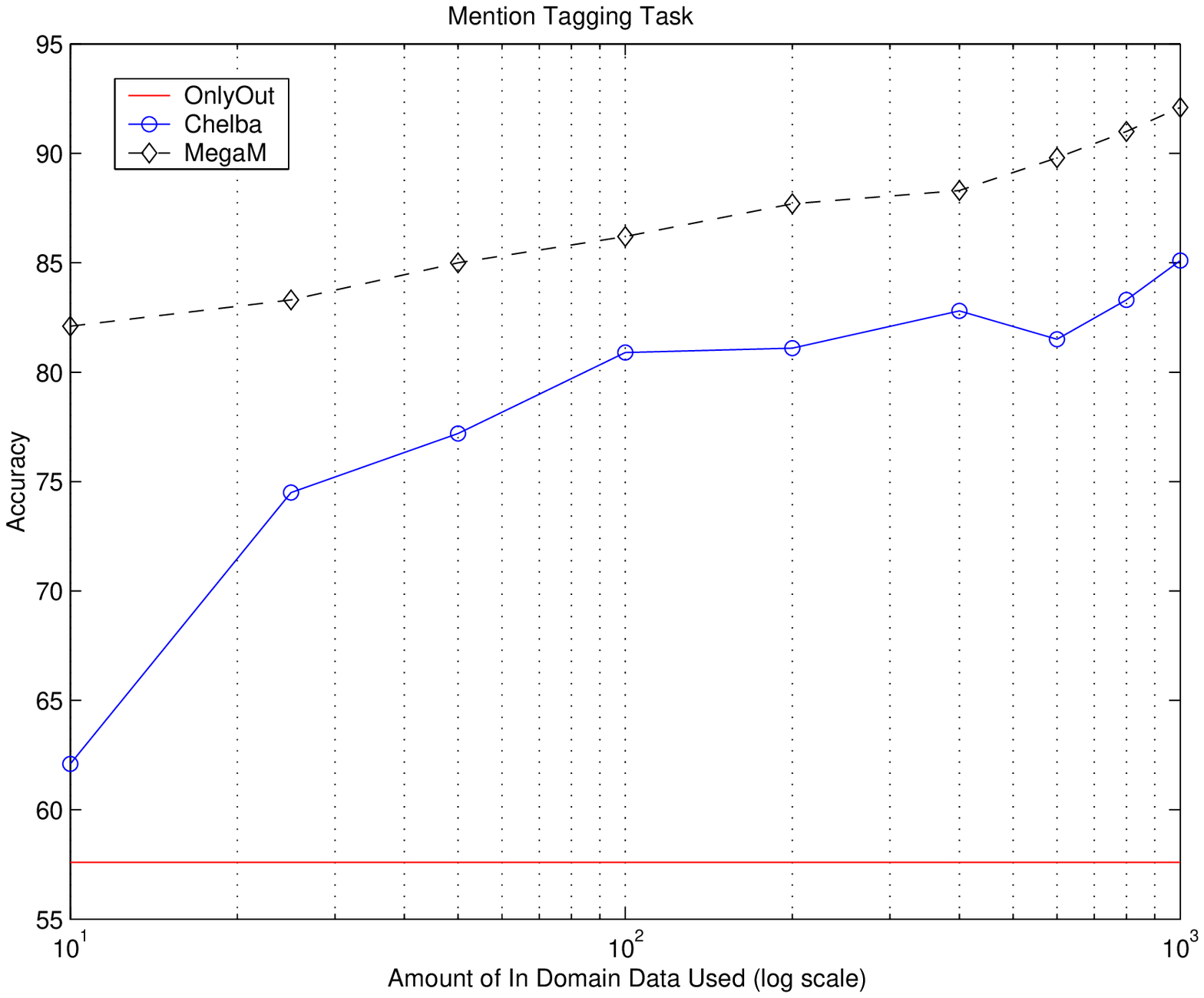,width=7.5cm}
\end{tabular}
}
\caption{Learning curves for {\bf\textsc{Prior}} and
  {\bf\textsc{MegaM}} models.}
\label{fig:curves}
\end{figure*}

Of particular interest is the amount of annotated \emph{in-domain}
data needed to see a marked improvement from the {\bf\textsc{OnlyO}}
baseline to a well adapted system.  We show in Figure~\ref{fig:curves}
the learning curves on the {\bf Mention Type} and {\bf Mention
Tagging} problems.  Along the $x$-axis, we plot the amount of
in-domain data used; along the $y$-axis, we plot the accuracy.  We
plot three lines: a flat line for the {\bf\textsc{OnlyO}} model that
does not use any in-domain data, and curves for the
{\bf\textsc{Prior}} and {\bf\textsc{MegaM}} models.  As we can see,
our model maintains an accuracy above both the other models, while the
Prior curve actually falls below the baseline in the type
identification task.\footnote{This is because the Fisher data is
personal conversations.  It hence has a much higher degree of first
and second person pronouns than news.  (The baseline that always
guesses ``person'' achieves a $77.8\%$ accuracy.)  By not being able
to intelligently use the out-of-domain data only when the in-domain
model is unsure, performance drops, as observed in the
{\bf\textsc{Prior}} model.}

\section{Model Introspection}

We have seen in the previous sections that the \megam\ routinely
outperforms competing models.  Despite this clear performance
improvement, a question remains open regarding the internal workings
of the models.  The $\pi\din$ variable captures the degree to which
the in-domain data set is truly in-domain.  The $z$ variables in the
model aim to capture, for each test data point, whether it is
``general domain'' or ``in-domain.''  In this section, we discuss the
particular values of the parameters the model learns for these
variables.

We present two analyses.  In the first (Section~\ref{sec:anal}), we
inspect the model's inner workings on the Mention Type task from
Section~\ref{sec:type}.  In this analysis, we look specifically at the
expected values of the hidden variables found by the model.  In the
second analysis (Section~\ref{sec:pi}), we look at the ability of the
model to judge degree of relatedness, as defined by the $\pi$
variables.

\subsection{Model Expectations} \label{sec:anal}

To focus our discussion, we will consider only the Mention Type task,
Section~\ref{sec:type}.  In Table~\ref{tab:performance}, we have shown
seven test-data examples from the Mention Type task.  The Pre-context
is the text that appears before the entity and the post-context is the
text that appears after.  We report the true class and the class our
model hypothesizes.  Finally, we report the probability of this
example being truly in-domain, according to our model.

\begin{table}[!t]
\center
\small
\begin{tabular}{|r@{\ \dots}c@{\dots\ }l|c|c|c|}
\hline
Pre-context & Entity & Post-context & True & Hyp & $\p{z=\textrm{I}}$ \\
\hline
my home is in trenton & new jersey & and that's where & GPE & GPE & 0.02 \\
veteran's administration & hospital & & ORG & LOC & 0.11 \\
you know by the american & government & because what is & ORG & ORG & 0.17 \\
gives & me & chills because if & PER & PER & 0.71 \\
is he capable of getting & anything & over here & WEA & PER & 0.92 \\
the fisher thing calling & me & ha ha they screwed up & PER & PER & 0.93 \\
when i was a & kid & that that was a & PER & PER & 0.98 \\
\hline
\end{tabular}
\caption{Examples from the test data for the Mention
Type task.  The ``True'' column is the correct entity type and the
``Hyp'' column is our model's prediction.  The final column is the
probability this example is truly in-domain under our model.}
\label{tab:performance}
\end{table}

As we can see, the three examples that the model thinks are general
domain are ``new jersey,'' ``hospital'' and ``government.''  It
believes that ``me,'' ``anything'' and ``kid'' are all in-domain.  In
general, the probabilities tend to be skewed toward $0$ and $1$, which
is not uncommon for na\"ive Bayes models.  We have shown two errors in
this data.  In the first, our model thinks that ``hospital'' is a
location when truly it is an organization.  This is a difficult
distinction to make: in the training data, hospitals were often used
as locations.

The second example error is ``anything'' in ``is he capable of getting
anything over here.''  The long-distance context of this example is a
discussion about biological warfare and Saddam Hussein, and
``anything'' is supposed to refer to a type of biological warhead.
Our model mistakingly thinks this is a person.  This error is likely
due to the fact that our model identifies that the word ``anything''
is likely to be truly in-domain (the word is not so common in
newswire).  It has also learned that most truly in-domain entities are
people.  Thus, lacking evidence otherwise, the model incorrectly
guesses that ``anything'' is a person.

It is interesting to observe that the model believes that the entity
``me'' in ``gives me chills'' is closer to general domain than
the ``me'' in ``the fisher thing calling me ha ha they screwed up.''
This likely occurs because the context ``ha ha'' has not occurred
anywhere in the out-of-domain training data, and twice in the
in-domain training data.  It is unlikely this example would have been
misclassified otherwise (``me'' is fairly clearly a person), but this
example shows that our model is able to take context into account in
deciding the domain.

All of the decisions made by the model, shown in
Table~\ref{tab:performance} seem qualitatively reasonable.  The
numbers are perhaps excessively skewed, but the ranking is believable.
The in-domain data is primarily from conversations about random (not
necessarily news worthy) topics, and is hence highly colloquial.
Contrastively, the out-of-domain data is from formal news.  The model
is able to learn that entities like ``new jersey'' and ``government''
have more to do with news that words like ``me'' and ``kid.''

\subsection{Degree of Relatedness} \label{sec:pi}

In this section, we analyze the values of $\pi$ found by the model.
Low values of $\pi\din$ and $\pi\dou$ mean that the in-domain data was
significantly different than the out-of-domain data; high values mean
that they were similar.  This is because a high value for $\pi$ means
that the general domain model will be used in most cases.  For all
tasks but Mention Type, the values of $\pi$ were middling around
$0.4$.  For Mention Type, $\pi\din$ was $0.14$ and $\pi\dou$ was
$0.11$, indicating that there was a significant difference between the
in-domain and out-of-domain data.  The exact values for all tasks are
shown in Table~\ref{tab:relatedness}.

\begin{table}[!t]
\center
\begin{tabular}{|l||c|c||c|c|}
\hline
          & {\bf Mention} & {\bf Mention} & {\bf Recap} & {\bf Recap}    \\
          & {\bf Type}    & {\bf Tagging} & {\bf CNN}   & {\bf ABC}      \\
\hline
$\pi\din$ & $0.14$        & $0.41$        & $0.36$      & $0.51$         \\
$\pi\dou$ & $0.11$        & $0.45$        & $0.40$      & $0.69$         \\
\hline
\end{tabular}
\caption{Values for the $\pi$ variables discovered by the \megam\ algorithm.}
\label{tab:relatedness}
\end{table}

These values for $\pi$ make intuitive sense.  The distinction between
conversation data and news data (for the Mention Type task) is
significantly stronger than the difference between manually and
automatically transcribed newswire (for the Mention Tagging task).
The values for $\pi$ reflect this qualitative distinction.  The rather
strong difference between the $\pi$ values for the recapitalization
tasks was not expected a priori.  However, a post hoc analysis shows
this result is reasonable.  We compute the KL divergence between a
unigram language model for the out-of-domain data set and each of the
in-domain data sets.  The KL divergence for the CNN data was $0.07$,
while the divergence for the ABC data $0.11$.  This confirms that the
ABC data is perhaps more different from the baseline out-of-domain
than the CNN data, as reflected by the $\pi$ values.

We are also interested in cases where there is little difference
between in-domain and out-of-domain data.  To simulate this case, we
have performed the following experiment.  We consider again the
Mention Type task, but use \emph{only} the training portion of the
out-of-domain data.  We randomly split the data in half, assigning
each half to ``in-domain'' and ``out-of-domain.''  In theory, the
model should learn that it may rely only on the general domain model.
We performed this experiment under ten fold cross-validation and found
that the average value of $\pi$ selected by the model was $0.94$.
While this is strictly less than one, it does show that the model is
able to identify that these are very similar domains.

\section{Conclusion and Discussion} \label{sec:conclusion}

In this paper, we have presented the \megam\ for domain adaptation in
the discriminative (conditional) learning framework.  We have
described efficient optimization algorithms based on the conditional
EM technique.  We have experimentally shown, in four data sets, that
our model outperforms a large number of baseline systems, including
the current state of the art model, and does so requiring
significantly less in-domain data.

Although we focused specifically on discriminative modeling in a
maximum entropy framework, we believe the novel, basic idea on which
this work is founded---to break the in-domain distribution $p\din$ and
out-of-domain distribution $p\dou$ into three distributions, $q\din$,
$q\dou$ and $q\dgg$---is general.  In particular, one could perform a
similar analysis in the case of generative models and obtain similar
algorithms (though in the case of a generative model, standard EM
could be used).  Such a model could be applied to domain adaptation in
language modeling or machine translation.

With the exception of the work described in Section~\ref{prior-work},
previous work in-domain adaptation is quite rare, especially in the
discriminative learning framework.  There is a substantial literature
in the language modeling/speech community, but most of the adaptation
with which they are concerned is based on adapting to new speakers
\cite{iyer97adaptlm,kalai99combining}.  From a learning perspective,
the \megam\ is most similar to a mixture of experts model.  Our model
can be seen as a \emph{constrained} experts model, with three experts,
where the constraints specify that in-domain data can only come from
one of two experts, and out-of-domain data can only come from one of
two experts (with a single expert overlapping between the two).  Most
attempts to build discriminative mixture of experts models make
heuristic approximations in order to perform the necessary
optimization \cite{jordan94experts}, rather than apply conditional EM,
which gives us strict guarantees that we monotonically increase the
data (incomplete) log likelihood of each iteration in training.

The domain adaptation problem is also closely related to multitask
learning (also known as learning to learn and inductive transfer).  In
multitask learning, one attempts to learn a function that solves many
machine learning problems simultaneously.  This related problem is
discussed by \citeA{thrun96nth}, \citeA{caruana97multitask} and
\citeA{baxter00inductive}, among others.  The similarity between
multitask learning and domain adaptation is that they both deal with
data drawn from related, but distinct distributions.  The primary
difference is that domain adaptation cares only about predicting one
label type, while multitask learning cares about predicting many.


As the various sub-communities of the natural language processing
family begin and continue to branch out into domains other than
newswire, the importance of developing models for new domains without
annotating much new data will become more and more important.  The
\megam\ is a first step toward being able to migrate simple
classification-style models (classifiers and maximum entropy Markov
models) across domains.  Continued research in the area of adaptation
is likely to benefit from other work done in active learning and in
learning with large amounts unannotated data.

\acks{We thank Ciprian Chelba and Alex Acero for making their data
available.  We thank Ryan McDonald for pointing out the \system{Feats}
baseline, which we had not previously considered.  We also thank Kevin
Knight and Dragos Munteanu for discussions related to this project.
This paper was greatly improved by suggestions from reviewers,
including reviewers of a previous, shorter version.  This work was
partially supported by DARPA-ITO grant N66001-00-1-9814, NSF grant
IIS-0097846, NSF grant IIS-0326276, and a USC Dean Fellowship to Hal
Daum\'e III.}

\appendix

\section{Conditional Expectation Maximization} \label{app:cem}

In this appendix, we derive Eq~\eqref{eq:Q} from
Eq~\eqref{eq:conditional} by making use of Jensen's inequality and the
variational bound.  The interested reader is referred to the work of
\citeA{jebara98cem} for further details.  Our discussion will consider
a bound in the \emph{change} of the log likelihood between iteration
$t-1$ and iteration $t$, $\De l^c$, as given in Eq~\eqref{eq:de-l-c}:

\pinch\pinch\pinch
\begin{align} \label{eq:de-l-c}
\De l^c 
 &= \log \frac {\p{y \| x ; \Th^t}} {\p{y \| x ; \Th^{t-1}}}
  = \log \frac {\p{x,y \| \Th^t} / \p{y \| \Th^t}} {\p{x,y \| \Th^{t-1}} / \p{y \| \Th^{t-1}}}
 \\
 &= \log \frac {\p{x,y ; \Th^t}} {\p{x,y ; \Th^{t-1}}}
  - \log \frac {\p{x   ; \Th^t}} {\p{x   ; \Th^{t-1}}}
  \label{eq:de-l-c2}
\end{align}

Here, we have effectively rewritten the log-change in the ratio of the
conditionals as the difference between the log-change in the ratio of
the joints and the log-change in the ratio of the marginals.  We may
rewrite Eq~\eqref{eq:de-l-c2} by introducing the hidden variables $z$
as:

\begin{equation} \label{eq:de-l-c-z}
\De l^c
  = \log \frac {\sum_z \p{x,y,z ; \Th^t}} {\sum_z \p{x,y,z ; \Th^{t-1}}}
  - \log \frac {\sum_z \p{x,z   ; \Th^t}} {\sum_z \p{x,z   ; \Th^{t-1}}}
\end{equation}

We can now apply Jensen's inequality to the first term in
Eq~\eqref{eq:de-l-c-z} to obtain:

\begin{equation}
\De l^c \geq
    \sum_z 
      \underbrace{
      \left[
      \frac {\p{x,y,z \| \Th^{t-1}}} {\sum_{z'} \p{x,y,z \| \Th^{t-1}}}
      \right]
      }_{h_{x,y,z,\Th^{t-1}}}
      \log \frac {\p{x,y,z ; \Th^t}} {\p{x,y,z ; \Th^{t-1}}}
  - \log \frac {\sum_z \p{x,z   ; \Th^t}} {\sum_z \p{x,z   ; \Th^{t-1}}}
  \label{eq:de-l-c-z2}
\end{equation}
\pinch\pinch\pinch

In Eq~\eqref{eq:de-l-c-z2}, the expression denoted
$h_{x,y,z,\Th^{t-1}}$ is the joint expectation of $z$ under the
previous iteration's parameter settings.  Unfortunately, we cannot
also apply Jensen's inequality to the remaining term in
Eq~\eqref{eq:de-l-c-z2} because it appears negated.  By applying the
variational dual ($\log x \leq x-1$) to this term, we obtain the
following, final bound:

\pinch\pinch\pinch
\begin{equation}
\De l^c \geq Q^t =
    \sum_z h_{x,y,z,\Th^{t-1}}
      \log \frac {\p{x,y,z ; \Th^t}} {\p{x,y,z ; \Th^{t-1}}}
  - \frac {\sum_z \p{x,z   ; \Th^t}} {\sum_z \p{x,z   ; \Th^{t-1}}}
  + 1
  \label{eq:de-l-c-z3}
\end{equation}

Applying the bound from Eq~\eqref{eq:de-l-c-z3} to the distributions
chosen in our model yields Eq~\eqref{eq:Q}.

\section{Derivation of Estimation Equations} \label{app:derivation}

Given the model structure and parameterization of the \megam given in
Section~\ref{subsec:megam}, Eq~\eqref{eq:hier-megam}, we obtain the
following expression for the joint probability of the data:

\pinch\pinch\pinch
\begin{eqnarray}
&&\hspace{-15mm}\p{\vec x, \vec y, \vec z
  \| \vec\ps\dnu, \vec\la\dnu, \pi} \nonumber\\
&=&
  \left\{\prod_{n=1}^N 
          \Ber(z_n \| \pi)
          \prod_{f=1}^F \Ber(x_{nf} \| \vec\ps^{z_n}_f)
          \textit{Gibbs}(y_n \| \vec x_n, \vec\la^{z_n})
        \right\}     \nonumber\\
&=&
  \left\{\prod_{n=1}^N
    \pi^{z_n} (1-\pi)^{1-z_n}
    \left[\prod_{f=1}^F
      \left(\vec\ps^{z_n}_f\right)^{x_{nf}}
      \left(1-\vec\ps^{z_n}_f\right)^{1-x_{nf}}\right]\right.\nonumber\\
&&\left.\hspace{30mm}
          \exp\left[\vec\la_{y_n}^{z_n} \T \vec x_n\right]
           \left(\sum_c \exp\left[\vec\la_c^{z_n} \T \vec x_n\right]\right)^{-1}
  \right\} \label{eq:joint}
\end{eqnarray}
\pinch\pinch\pinch

The marginal distribution is obtained by removing the last two terms
(the $\exp$ and the sum of $\exp$s) from the final equation.  Plugging
Eq~\eqref{eq:joint} into Eq~\eqref{eq:Q} and using the notation from
Eq~\eqref{eq:notation}, we obtain the following expression for $Q^t$:

\pinch\pinch\pinch
\begin{eqnarray}
Q^t &=&
  \sum_\nu \left[ 
    \log \Nor(\vec\la\dnu; 0, \si^2 I) +
    \sum_{f=1}^F \log \Bet(\vec\ps\dnu_f; a,b)
    \right] \nonumber\\
&+&
  \sum_{n=1}^N \Bigg[
    \sum_{z_n} h_n \Bigg\{
        z_n \log \pi + (1-z_n) \log (1-\pi)
      + \log \ps_{n,z_n} \nonumber\\
&&  \hspace{25mm}
      + \sum_{f=1}^F x_{nf} \log \vec\la_{y_n}^{z_n}
      - \log \sum_c \exp\left[\vec\la_c^{z_n} \T x_n\right]
      - j_{n,z_n}^t
      \Bigg\} \nonumber\\
&&  \hspace{15mm}
    - m_n^{t-1} \pi^{z_n} (1-\pi)^{1-z_n}
      \ps_{n,z_n}
    + 1 \Bigg]  \label{eq:megaQ}
\end{eqnarray}
\pinch\pinch\pinch

\noindent as well as an analogous term for the out-of-domain data.
$j$ and $m$ are defined in Table~\ref{tab:notation}.

\subsection{M-Step for $\pi$}

For computing $\pi$, we simply differentiate $Q^t$ (see
Eq~\eqref{eq:megaQ}) with respect to $\pi$, obtaining:

\begin{equation}
\frac {\partial Q^t} {\partial \pi}
= \sum_{n=1}^N \frac {h_n} \pi      +
                 \frac {1-h_n} {1-\pi}  +
    m_n^{t-1} \left(\ps_{n,0} - \ps_{n,1}\right)
\end{equation}

\noindent
solving this for $0$ leads directly to a quadratic expression of the
form:

\begin{eqnarray}
0 &=& \pi^2 \left[\sum_{n=1}^N m_n^{t-1} \left(\ps_{n,0} - \ps_{n,1}\right)\right]
  \nonumber\\
  &+& \pi^1 \left[-1 + \sum_{n=1}^N 
        \(2 h_n - m_n^{t-1} \left(\ps_{n,0} - \ps_{n,1}\right)\)\right]
  \nonumber\\
  &+& \pi^0 \left[-\sum_{n=1}^N h_n \right]
  \label{eq:mstep-pi}
\end{eqnarray}

Solving this directly for $\pi$ gives the desired update equation.

\subsection{M-Step for $\vec\la$}

For optimizing $\vec\la\din$, we rewrite $Q^t$, Eq~\eqref{eq:megaQ},
neglecting all irrelevant terms, as:

\begin{equation}
Q^t[\vec\la]
=  \sum_{n=1}^{N} \(1-h_n\) \left\{
    \sum_{f=1}^{F} x_{nf} \vec\la_{y_n,f} -
    \log \sum_c \exp\left[\vec\la_c \T x_n\right]\right\}
  + \log \Nor(\vec\la; 0, \si^2 I)
   \label{eq:mstep-phi-1}
\end{equation}

In Eq~\eqref{eq:mstep-phi-1}, the bracketed expression is exactly the
log-likelihood term obtained for standard logistic regression models.
Thus, the optimization of $Q$ with respect to $\vec\la\din$ and
$\vec\la\dou$ can be performed using a weighted version of standard
logistic regression optimization, with weights defined by $(1-h_n)$.
In the case of $\vec\la\dgg$, we obtain a weighted logistic regression
model, but over all $N\din+N\dou$ data points, and with weights
defined by $h_n$.

\subsection{M-Step for $\vec\ps$}

In the case of $\vec\ps\din$ and $\vec\ps\dou$, we rewrite
Eq~\eqref{eq:megaQ} and remove all irrelevant terms, as:

\begin{equation}
Q^t[\vec\ps\din]
= \sum_{f=1}^F \log \Bet(\ps_f; a,b) +
  \sum_{n=1}^N  \left[
    \(1-h_n\) \log \ps_{n,0}
    - m_n^{t-1} (1-\pi) \ps_{n,0} \right]
\end{equation}

\noindent
Due to the presence of the product term in $\vec\ps$, we cannot
compute an analytical solution to this maximization problem.  However,
we can take derivatives component-wise (in $F$) and obtain analytical
solutions (when combined with the prior).  This admits an iterative
solution for maximizing $Q^t_{\vec\ps}$ by maximizing each component
separately until convergence.  Computing derivatives of $Q^t$ with
respect to $\ps_f$ requires differentiating $\ps_{n,0}$ with respect
to $\ps_f$; this has a convenient form (recalling the notation from
Table~\ref{tab:notation}:

\begin{equation}
\frac \partial {\partial \ps_f}
   \ps_{n,0}
= \left[\ps_{n,0,-f}\right]
  \frac \partial {\partial \ps_f}
    \left\{ x_{nf}\ps_f + (1-x_{nf})(1-\ps_f) \right\}
= \ps_{n,0,-f}
\end{equation}

Using this result, we can maximize $Q^t$ with respect to $\ps_f$ by
solving:

\begin{eqnarray}
\frac \partial {\partial \ps_f} \left[
  Q^t_{\ps_f} \right]
&=& 
  \sum_{n=1}^N \Bigg[
  \(1-h_n\) \frac {
          x_{nf} (1-\ps_f) -
                  (1-x_{nf}) \ps_f}
        { \ps_f \(1-\ps_f\) } \label{eq:qbet00}\\
&&\hspace{20mm}- j_{n,0}(1-\pi) \ps_{n,0,-f} \Bigg]
  + \frac 1 { \ps_f \(1-\ps_f\) } \nonumber\\
&=& 
  \frac 1 { \ps_f \(1-\ps_f\) }
    \left[ 1 + \sum_{n=1}^N \(1-h_n\) (x_{nf} - \ps_f) \right]
  - \sum_{n=1}^N j_{n,0}(1-\pi) \ps_{n,0,-f} \nonumber
\end{eqnarray}

Equating this to zero yields a quadratic expression of the form:

\begin{eqnarray}
0 
&=&
\left(\ps_f\right)^2 
  \left[\sum_{n=1}^N j_{n,0} (1-\pi) \ps_{n,0,-f}\right] \nonumber\\
&+&
\left(\ps_f\right)^1
  \left[-\sum_{n=1}^N \Big(1-h_n + j_{n,0}(1-\pi) \ps_{n,0,-f}\Big)
    \right] \nonumber\\
&+&
\left(\ps_f\right)^0
\left[1 + \sum_{n=1}^N \(1-h_n\) x_{nf}\right]
\label{eq:mstep-psi-beta00}
\end{eqnarray}

This final equation can be solved analytically.  A similar expression
arises for $\ps\dou_f$.  In the case of $\ps\dgg_f$, we obtain
a quadratic form with sums over the entire data set and with $h_n$
replacing the occurrences of $(1-h_n)$:



\begin{eqnarray}
0 
&=&
\left(\ps\dgg_f\right)^2 
  \left[\sum_{n=1}^{N\din} j\din_{n,1} \pi\din \ps\din_{n,1,-f}
      + \sum_{n=1}^{N\dou} j\dou_{n,1} \pi\dou \ps\dou_{n,1,-f}
    \right] \nonumber\\
&+&
\left(\ps\dgg_f\right)^1
  \Bigg[-\sum_{n=1}^{N\din} \Big(h\din_n + j\din_{n,1}\pi\din \ps\din_{n,1,-f}\Big)
        -\sum_{n=1}^{N\dou} \Big(h\dou_n + j\dou_{n,1}\pi\dou \ps\dou_{n,1,-f}\Big)
    \Bigg] \nonumber\\
&+&
\left(\ps\dgg_f\right)^0
\left[1 + \sum_{n=1}^{N\din} h\din_n x\din_{nf} + 
          \sum_{n=1}^{N\dou} h\dou_n x\dou_{nf}
\right]
\label{eq:mstep-psi-beta00b}
\end{eqnarray}

\noindent Again, this can be solved analytically.  The values $j$,
$m$, $\ps_{\cdot,\cdot}$ and $\ps_{\cdot,\cdot,-\cdot}$ are defined in
Table~\ref{tab:notation}.

\vskip 0.2in

\end{document}